\documentclass[runningheads]{llncs}
\usepackage{graphicx}
\usepackage{soul}
\usepackage{xcolor}
\usepackage{url}
\begin{document}
\title{Leveraging Weakly Annotated Data for Hate Speech Detection in Code-Mixed Hinglish: A Feasibility-Driven Transfer Learning Approach with Large Language Models}
\author{Sargam Yadav\inst{1}\orcidID{0000-0001-8115-6741} \and
Abhishek Kaushik\inst{2}\orcidID{0000-0002-3329-1807} \and
Kevin McDaid\inst{3}\orcidID{0000-0002-0695-9082}}
\authorrunning{S. Yadav et al.}
\institute{Dundalk Institute of Technology, Dundalk, A91 K584, Ireland \\ \email{sargam.yadav@dkit.ie} \and Dundalk Institute of Technology, Dundalk, A91 K584, Ireland
\email{abhishek.kaushik@dkit.ie} \and Dundalk Institute of Technology, Dundalk, A91 K584, Ireland \email{kevin.mcdaid@dkit.ie}}
\maketitle              % typeset the header of the contribution
\begin{abstract}
The advent of Large Language Models (LLMs) has advanced the benchmark in various Natural Language Processing (NLP) tasks. However, large amounts of labelled training data are required to train LLMs. Furthermore, data annotation and training are computationally expensive and time-consuming. Zero and few-shot learning have recently emerged as viable options for labelling data using large pre-trained models. Hate speech detection in mix-code low-resource languages is an active problem area where the use of LLMs has proven beneficial. In this study, we have compiled a dataset of 100 YouTube comments, and weakly labelled them for coarse and fine-grained misogyny classification in mix-code Hinglish. Weak annotation was applied due to the labor-intensive annotation process. Zero-shot learning, one-shot learning, and few-shot learning and prompting approaches have then been applied to assign labels to the comments and compare them to human-assigned labels. Out of all the approaches, zero-shot classification using the Bidirectional Auto-Regressive Transformers (BART) large model and few-shot prompting using Generative Pre-trained Transformer- 3 (ChatGPT-3) achieve the best results\footnote{This paper is accepted in the 16th ISDSI-Global Conference 2023 https://isdsi2023.iimranchi.ac.in/}.

\end{abstract}
\section{Introduction}

The infiltration of the internet throughout the world has supported the rise in popularity of social media platforms. Online platforms have transformed the way several daily tasks, such as business operations and communication, are performed. Although social media has a list of potential benefits, it also allows individuals to spread misinformation, engage in cyberbullying \cite{moreno2019applying}, and make hateful statements online \cite{kumar2018benchmarking}. These actions can greatly impact the well-being of the targets. Machine Learning (ML) and Deep Learning (DL) models have shown great potential in performing content moderation \cite{mandl2020overview} \cite{fersini2020ami}, even on low-resource and code-switched languages. Code-switched languages combine features of two languages and are used in multilingual societies. Despite the progress, there is still room for significant improvement, to ensure that poorly trained models do not restrict free speech \cite{yadav2023comprehensive}.

Zero-shot, one-shot, and few-shot learning are alternate approaches for performing text classification when labelled data is scarce \cite{yin2019benchmarking}. Zero-shot learning is performed by using a model to classify instances from previously unseen classes without any previous training \cite{del2023respectful}. In one-shot learning, one example from each class is provided to the model for training and unseen additional examples are used to evaluate the model \cite{yan2018few}. In few-shot learning, a few examples are previously seen during training for each class and evaluation is performed using unseen examples \cite{lin2022few}. These approaches can help reduce the number of labelled instances used to train a hate speech classifier. This can be beneficial in low-resource and code-mixed languages where there is a scarcity of data.

%\textcolor{blue}{please how some example of few short learning and zero shot learning with images}

In this study, we have implemented and evaluated zero-shot, one-shot, and few-shot text classification techniques on a weakly annotated dataset consisting of misogynistic and non-misogynistic YouTube comments. We wanted to examine the benefits of transfer leaning approaches on this small dataset to explore alternatives to large curated datasets. The data was collected using YouTube API from a video titled `What people think about bringing the Marital rape law in India?'\footnote{https://www.youtube.com/watch?v=UBcOqt17FgI}. Details are discussed in Section \ref{dataset}. Ethical permission for collecting the data has been obtained from the host institute. % The data collected process was started after obtaining ethical permission from the research committee at Dundalk Institute of Technology. 

\begin{figure}[!ht]
\includegraphics[width=\textwidth]{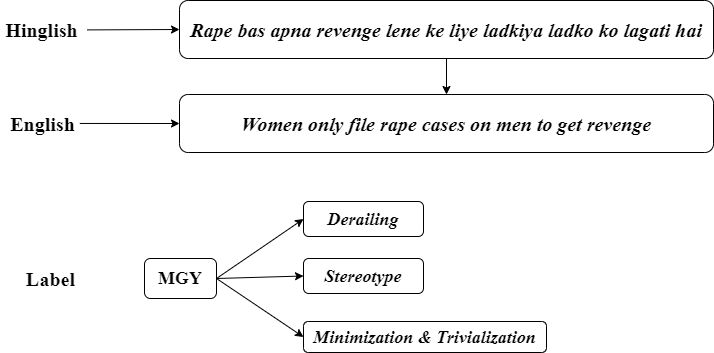}
\caption{Example of a sentence from the dataset belonging to class MGY} \label{Example}
\end{figure}

The comments are first divided into 2 classes: misogynistic (MGY) and non-misogynistic (NOT). Fig. \ref{Example} shows an example from the dataset and its corresponding class and labels. Then, the comments that were positive for the class `MGY' were further labelled into 9 different labels depicting the type of misogyny. The 9 fine-grained labels are `Derailing', `Sexual Harassment and Threats of Violence', `Stereotyping', `Minimization and Trivialization', `Religion-based', `Whataboutism'. `Shaming', `Moral Policing', and `Victim Blaming'. A particular sample from the `MGY' class could belong to more than one label in fine-grained classification. Zero-shot and few-shot models are trained and evaluated on the dataset. Table \ref{tab:multi_label} gives the count and description of each fine-grained label. ChatGPT-3 \cite{chatgpt} has also been used to perform few-shot prompting.

\begin{table}[!ht]
    \centering
    \begin{tabular}{p{3cm}p{2cm}p{8cm}}
    \hline
    \textbf{Label Name}     & \textbf{Label Count}  & \textbf{Label Description} \\
    \hline
     Derailing & 23 & Targeting slurs towards women without any broader intentions\\
   Sexual Harassment and Threats of Violence & 1 & Exertion of physical dominance over or intimidating them using threats \\
    Stereotyping & 9& Comparing women to narrow standards\\
Minimization and Trivialization &11 &  Downplaying the problem and making it seem insignificant \\
    Religion based & 1 & Sexist discrimination or prejudice that arise from religious scriptures \\
Whataboutism & 8 & Using rhetorical tactic to deflect criticism\\
Shaming & 4 & Inappropriate comments made about women deviating from conservative expectations\\
Moral Policing & 2 & enforcing perceived morals regarding factors such as clothing, relationships, etc\\
Victim Blaming & 2 & Holding the victim of wrongful act entirely or partially at fault\\

         \hline
    \end{tabular}
    \caption{Multi-label categorization of the dataset}
    \label{tab:multi_label}
\end{table}

The rest of the article is structured as follows: Section \ref{mot} describes the motivation behind conducting the study, the research hypothesis, and research questions. Section \ref{lit} discusses previous work in the fields of hate speech detection, zero and few-shot learning, aiming to draw insights from it. Section \ref{methodology} details the dataset and experimental procedure used in the study. Section \ref{results} reports the results of the experiment and analysis of the results. Section \ref{discuss} discusses the findings of the study with respect to the research questions and hypothesis. In Section \ref{limitations}, the limitations of the study are addressed. Section \ref{conc} provides a conclusion to the study. 

\section{Motivation} \label{mot}

Hate speech has become rampant in today's online discourse. This type of targeted hate can have a detrimental impact on the target, silencing their online voice. Artificial intelligence models, such as Bidirectional Encoder Representations from Transformers (BERT) \cite{devlin2018bert}, Cross-lingual Language Model (XLM-RoBERTa) \cite{conneau2019unsupervised}, ChatGPT-3, etc., can be trained and evaluated on annotated datasets, and have proven to be useful in combating much of the offensive and hateful comments online \cite{mandl2020overview} \cite{fersini2020ami} \cite{fersini2018ibereval}. However, large and complex deep learning models require large amounts of data for training. Also, the training process is time-consuming and computationally expensive. The annotation of the data also requires a lot of human effort \cite{basile2019semeval}. Recent studies have shown that LLMs that have been pre-trained on large amounts of data in an unsupervised manner inherently store some understanding of the language and can be used to perform zero and few-shot classification tasks \cite{brown2020language} \cite{del2023respectful}. This study tests these finding on a new Hinglish dataset that consists of fine-grained and coarse-grained classification of misogynistic comments. Hinglish is an informal combination of Hindi and English, where Hindi characters are Romanised and intermixed with English words \cite{mathur2018did}. Code-mixed languages present unique challenges due to lack of consistent spelling, grammatical structure, and syntax \cite{mathur2018did}. 

The research hypothesis explored in the study is as follows:

\textit{Hypothesis: Zero-shot, one-shot, and few-shot learning can be reliably used to assign labels to mix-code Hinglish YouTube comments for coarse and fine-grained misogyny classification.}

The following research questions explore our research hypothesis:

\begin{enumerate}
    \item How does zero-shot learning perform at the task of labelling YouTube comments for coarse and fine-grained misogyny classification in mix-code Hinglish?
    
    \item How does one-shot and few-shot learning perform at the task of labelling YouTube comments for coarse and fine-grained misogyny classification in mix-code Hinglish?
\end{enumerate}

\section{Literature Review} \label{lit}

In this section, we will discuss the recent state-of-the-art approaches for hate speech classification, as well as zero-shot, one-shot, and few-shot learning models used in NLP.

\subsection{Hate Speech Classifiers and Code-mixed Languages}

Hate speech detection is an active area of study. Over the past few years, there has been an increase in the number of studies tackling hate speech detection in English and other languages such as Italian \cite{fersini2020ami}, Bengali \cite{kumar2020evaluating}, Hindi \cite{mandl2020overview}, mix-code Hinglish, and more \cite{satapara2021overview}. Various shared tasks and challenges are hosted regularly in several languages to advance research in the area. For example,  Hate Speech and Offensive Content Identification in English and Indo-Aryan Languages (HASOC) is conducted in several languages such as English, Hindi, Marathi, mix-code Hinglish, and more \cite{mandl2020overview}.
%The Automatic Misogyny Detection (AMI)  \cite{fersini2020ami} \cite{fersini2018ibereval} was conducted for building classifiers for detecting the presence and type of misogyny in the text in English, Italian, and Spanish. Transformer-based models, such as the BERT, XLM-RoBERTa, IndicBERT \cite{kakwani2020indicnlpsuite}, etc. are the current state of the art, but they do not perform too well with fine-grained misogyny classification.

There has been a significant effort in studying Hinglish for various NLP tasks. One popular Hinglish dataset is the `YouTube cookery channels' dataset consisting of Hinglish and English comments \cite{datasetkaur}. Several studies have been performed on this dataset for sentiment classification through different approaches, such as TF-IDF vectorizer and logistic regression model \cite{kaur2019cooking}, multi-layer perceptron \cite{donthula2019man}, fine-tuning BERT-based models \cite{yadav2021cooking} and utilizing contextual embeddings from BERT-based models as features for downstream tasks \cite{yadav2022contextualized}.

\subsection{Zero-Shot, One-shot, and Few-Shot Learning}

Several recent studies on LLMs have suggested that the final fine-tuning step can be removed, and the models can function as few-shot classifiers \cite{radford2019language} \cite{brown2020language}. Models such as GPT-3 have unique generation capabilities that can be leveraged for specific tasks. For example, Yin et al. \cite{yin2019benchmarking} define zero-shot classification as a textual entailment framework. They propose using pre-trained Natural Language Inference (NLI) models, such as BART \cite{lewis2019bart} and RoBERTa as zero-shot classifiers. The comment is posed as a premise and a hypothesis is constructed for each label. It gives excellent performance in mapping corrupted documents to the original. Languages models such as ChatGPT-3 show that few-shot prompting is an efficient strategy that can be used for text classification \cite{beltagy2022zero}. Table \ref{tab:zero_one} provides a distinction between zero-shot, one-shot, and few-shot learning.

\begin{table}[!ht]
    \centering
    \begin{tabular}{cc}
    \hline
        \textbf{Zero-Shot} & No labelled training data used   \\
        \textbf{One-Shot} & One sample comment of each class is used for training\\
        \textbf{Few-Shot} & Few samples from each class used for training\\
        \hline
    \end{tabular}
    \caption{Distinction between Approaches Used in the Study}
    \label{tab:zero_one}
\end{table}

Plaza-del-Arco et al. \cite{del2023respectful} performed a similar study where they performed zero-shot classification on hate speech datasets using the following models: BERT, RoBERTa and DeBERTa. The findings suggested that with proper instruction fine-tuning, zero-shot classification can compete with the performance of fine-tuned models. Studies for multilingual and low-resource data have also been performed using few-shot learning, and observe that prompting can achieve good results \cite{mozafari2022cross} \cite{lin2022few}. For example, few-shot learning through GPT-3 achieved 85\% results for hate speech detection \cite{chiu2021detecting}. 
SetFit \cite{tunstall2022efficient} is a framework based on fine-tuning Sentence Transformers and can achieve results that are competitive with models that are fine-tuned on large amounts of data. It does this by generating rich embeddings from a small sample of data. It can be trained without any prompts, is faster, and supports multilingual text. In this study, the English version of the Masked and Permuted Pre-training for Language Understanding (MPNET) \cite{song2020mpnet} model is used. SetFit first fine-tunes the model on the small number of samples, then trains a classification head on the generated embeddings.

Thus, it becomes apparent that training and deploying complex deep learning architectures for hate speech classification is a labour and cost-intensive process. The availability of large amounts of labelled data is also a hurdle, especially in low-resource and mix-code languages. Zero-shot and few-shot learning can provide an opportunity to create toolkits that can work with scarce amounts of data.

\section{Methodology}\label{methodology}

\begin{figure}
 \centering
\includegraphics[width=0.8\textwidth]{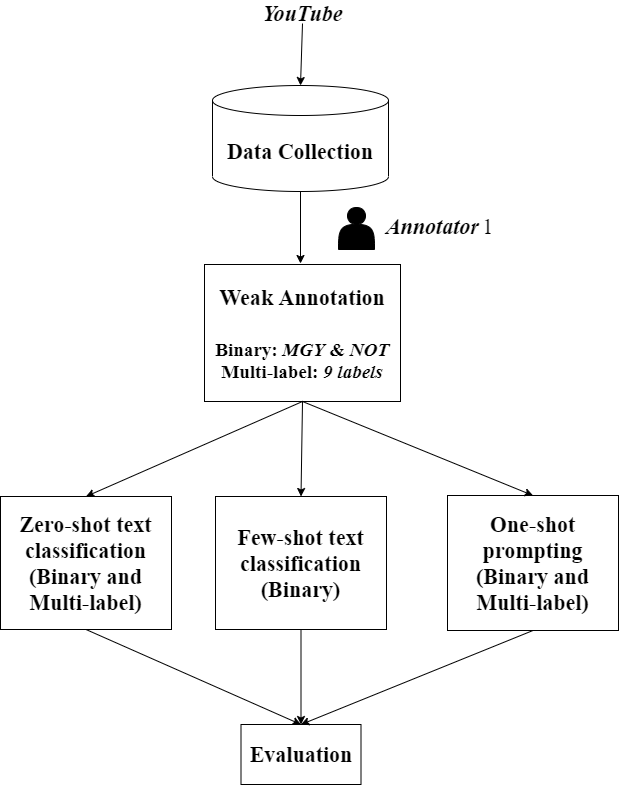}
\caption{Methodology of the proposed study} \label{method}
\end{figure}

In this section, we have discussed the details of the dataset used in the study and the experimental setup. Fig \ref{method} displays a flowchart of the experimental methodology.

\subsection{Dataset}
\label{dataset}

The dataset used in this study has been extracted from YouTube videos about current events in India. The dataset consists of 100 comments in both English and code-mixed Hinglish. The comments have been weakly annotated by a single annotator at two levels. Firstly, the comment is classified as either `MGY' (misogynistic) or `NOT' (Not misogynistic). There are a total of 75 comments belonging to the class `NOT' and 25 comments of the class `MGY'. Secondly, instances belonging to the class `MGY' were further labelled into 9 categories, where each instance can belong to more than category. The 9 categories were inspired by the work of Anzovino et al. \cite{anzovino2018automatic} and Abburi et al. \cite{abburi2021fine}. They are as follows: Derailing, Sexual Harassment and Threats of Violence, Stereotyping, Minimization and Trivialization, Religion based, Whataboutism, Shaming, Moral Policing, and Victim Blaming. The dataset is severely imbalanced, with only 25 positive instances for the `MGY' class. This represents real-world scenarios, where hateful comments are scarce. Ideally, the dataset should be annotated by at least two annotators and inter-annotator agreement should be calculated using the kappa coefficient \cite{kraemer2014kappa}. The dataset is currently in the process of being annotated by a second annotator, and kappa coefficient will be calculated accordingly.

%\begin{table}[!ht]
 %   \centering
 %   \begin{tabular}{c|c}
%    \hline
%    \textbf{Label Name}     & \textbf{Label Count} \\
 %   \hline
 %      MGY  &  73\\
%      NOT   & 27 \\
 %     \textbf{Total}   & \textbf{100} \\
%         \hline
%    \end{tabular}
 %   \caption{Binary categorization of the dataset}
 %   \label{tab:binary_label}
%\end{table}

\subsection{Experiment}

In this section, we will describe the experimental methodology used in the study.

\subsubsection{Zero-shot Learning}

\begin{figure}
    \centering
    \includegraphics[width=1.0\textwidth]{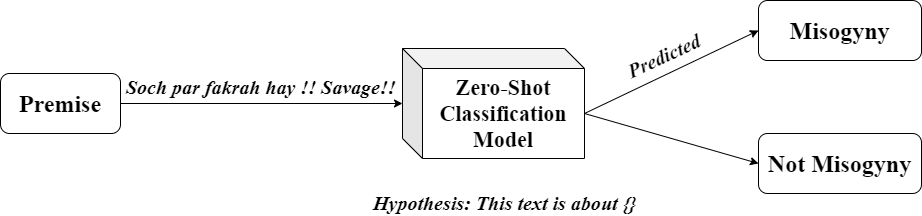}
    \caption{Zero-Shot Classification}
    \label{fig:zeroshot}
\end{figure}

In this study, the BART large model is implemented using HuggingFace\footnote{https://huggingface.co/tasks/zero-shot-classification}. The BART large model is trained on the MNLI dataset with 407 million parameters\footnote{https://huggingface.co/facebook/bart-large-mnli}. It functions as a ready-made zero-shot classifier. According to Yin et al. \cite{yin2019benchmarking}, we can consider the problem of zero-shot classification as an Natural Language Inference (NLI) problem, where the sentence we want to classify is treated as the premise, and the two labels (MGY or NOT) are treated as the hypothesis. The model will predict if the premise entails the hypothesis. For example, if we want to check if a statement belongs to the `MGY', the hypothesis would be constructed as \textit{`This text is about misogyny'}. The model consists of 12 encoder and decoder layers and 407 million parameters \cite{tunstall2022efficient}. Each sequence is predicted independently and compared with the corresponding annotated value. This procedure is repeated for both binary classification and multi-label classification. Fig \ref{fig:zeroshot} describes the process of performing zero-shot classification on an example sequence.

%Accuracy, precision, recall, macro-f1, and MCC are reported for the binary classification task. For multi-label classification, only precision, recall and f1 scores weighted by samples are calculated. 

\subsubsection{Few-shot Learning}

To perform one-shot and few-shot binary classification, the English MPNet (`paraphrase-mpnet-base-v2')\footnote{https://huggingface.co/sentence-transformers/paraphrase-mpnet-base-v2} sentence transformer is implemented through HuggingFace using the SetFit\footnote{https://huggingface.co/blog/setfit} approach. For one-shot learning, one example of each class was given as training data. For few-shot classification, four samples of each class were used to train the model. The model was evaluated on the remaining samples of the dataset. 

%Scores for accuracy, macro-f1, precision, recall, and MCC and calculated for comparison.

\subsubsection{One-shot prompting through ChatGPT-3}

The aim of few-shot promoting using ChatGPT-3\footnote{https://chat.openai.com/} in this study is to verify if the model can correctly label comments. The model has been accessed using its website. GPT-3 uses the standard transformer architecture with only encoder blocks and is trained on 570 GB of text data \cite{brown2020language}. It consists of 96 attention blocks with 96 attention heads each. ChatGPT-3 \cite{chatgpt} is a variant of GPT-3, which was modified by supervised fine-tuning of GPT-3.5, a reinforcement learning-based reward model, and Proximal Policy Optimization (PPO). ChatGPT-3 is powered by the `gpt-3.5-turbo' model.

%\begin{figure}
%\includegraphics[width=\textwidth]{figures/example.png}
%\caption{Taxonomy} \label{taxonomy}
%\end{figure}

\section{Results and Analysis} \label{results}

In this section, we will discuss the results of the experiment and analyse the findings.

\subsection{Zero-Shot, One-Shot, and Few-shot Classification} \label{zero}

For measuring the performance of zero-shot binary and multi-label classification, the classes and labels assigned by the annotator are treated as the base values and the results given by the model are treated as the predicted value. 

Table \ref{tab:few_shot} displays the results of the zero-shot, one-shot, and few-shot models for the coarse-grained misogyny classification. For zero-shot classification, the models are used with its default parameters and no additional fine-tuning is performed. The highest accuracy achieved is 54\%, which is not very high. Precision, recall, and F1-score are calculated as a macro-average. 

\begin{figure}
    \centering
    \includegraphics[width=0.9\textwidth]{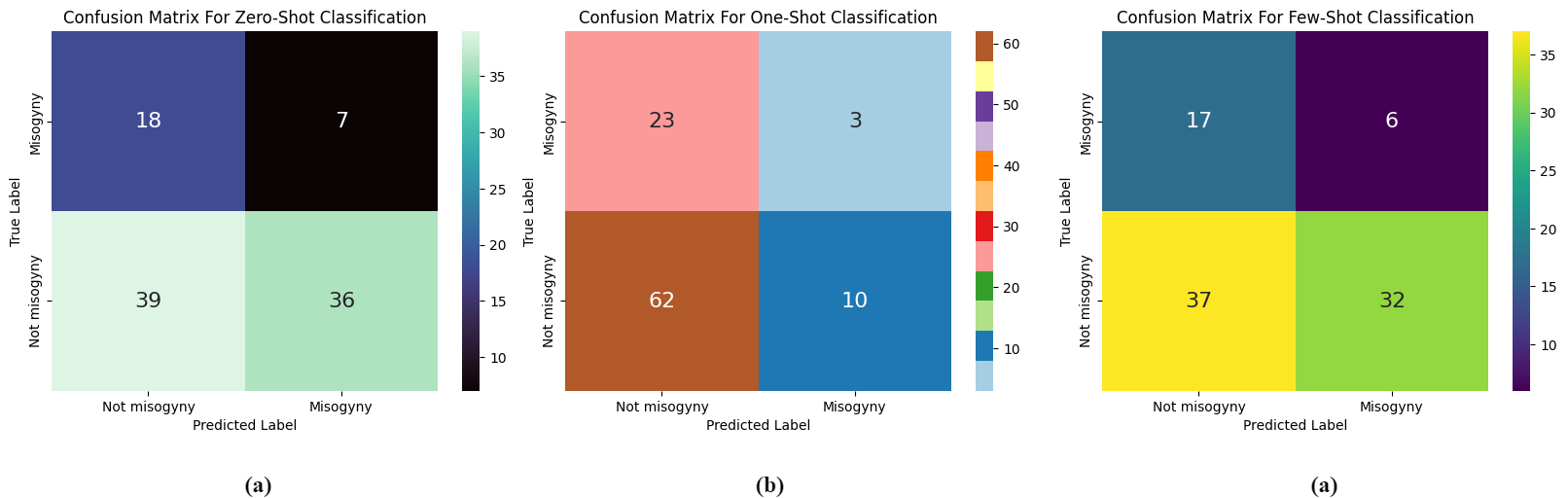}
    \caption{Confusion Matrix for Classifiers}
    \label{fig:conf}
\end{figure}

\begin{table}[!ht]
    \centering
    \begin{tabular}{cccccc}
    \textbf{Approach} & \textbf{Accuracy} & \textbf{Macro-F1} & \textbf{Precision} & \textbf{Recall} & \textbf{MCC} \\
    \hline
    Zero-shot & 0.54 & 0.5245&  0.5764 & 0.6 & 0.5245\\
      One-shot   &  0.3367 & 0.3248 & 0.5199 & 0.5117 & 0.0305  \\
        Few-shot & 0.5326 & 0.5198 & 0.5784 & 0.6014 & 0.1784 \\
        \hline
    \end{tabular}
    \caption{Results of Zero-shot, One-shot, and Few-shot classification}
    \label{tab:few_shot}
\end{table}
Fig \ref{fig:conf} displays the confusion matrix for the zero-shot, one-shot, and few-shot classification models. For the zero-shot classifier, the true positives for the class `MGY' is 18, out of 25 total samples. However, for the class `NOT', the number of true positives is only 39 out of 75 total samples.  

%\textcolor{blue}{talk about the impact about confusion metrix}

The results for fine-grained multi-label misogyny classification are as follows: precision = 0.2759. recall = 0.2314, and f1  = 0.2222. Precision, recall and F1 score are weighted by the number of samples. The precision, recall, and F1-scores are very low, indicating that zero-shot learning does not perform very well at the task of multi-label misogyny classification, at least for a small dataset of this size. The model seems to misclassify statements containing words such as `rape' and `sex'. For example, the statement `Mr sehbaz Jiyada meretial rapes to tumhare me hota h', which translates to `There are more marital rapes in your culture.' This statement is incorrectly classified as misogynistic. 

%\begin{table}[!ht]
 %   \centering
 %   \begin{tabular}{c|c}
  %  \hline
   % \textbf{Metric}     & \textbf{Score} \\
    %\hline
       
    % Precision & 0.2759\\
    % Recall & 0.2314\\
     %F1  & 0.2222\\
      
     %    \hline
    %\end{tabular}
    %\caption{Results of Zero-Shot Multi-Label classification}
    %\label{tab:results_multi}
%\end{table}

\begin{figure}
    \centering
    \includegraphics[width=0.9\textwidth]{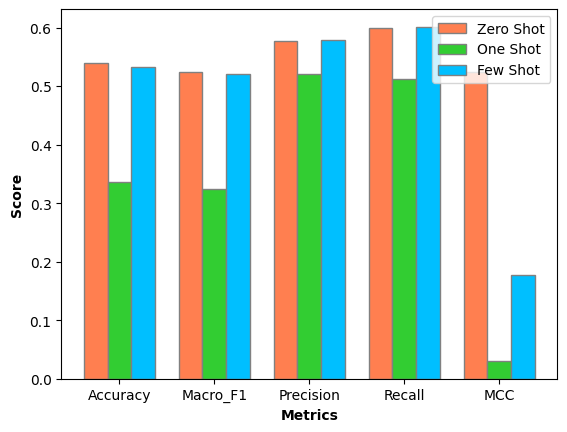}
    \caption{Bar Graph for Binary Classification}
    \label{fig:bar-graph}
\end{figure}

Out of the three approaches for binary classification, zero-shot classification provides the highest accuracy of 54\%, followed by 51\% achieved by few-shot learning. Fig \ref{fig:bar-graph} depicts the performance of one-shot, few-shot, and zero-shot learning models on all metrics. 

\subsection{One-shot Prompting}

ChatGPT-3 is a natural language generation model, which can process up to 3125 words at a time. It has been used to perform few-shot prompting. In this study, the default parameters of the model have not been updated.

The following two prompts have been tested on the model.
\begin{enumerate}
    \item Is this statement in mix-code Hinglish misogynistic? ``Delhi me randi jada hai ish liye rape bhi jada hai". This translates to `There are more prostitutes in Delhi, therefore rape is also more' in English.
    \item Which labels can be applied to the statement out of the following: Derailing, Sexual Harassment \& Threats of Violence, Stereotype, Minimization \& Trivialization,	Religion-based,	What aboutism, Shaming,  Moral Policing, and  Victim Blaming?
\end{enumerate}

%Figure \ref{1} depicts the result of the first prompt given by the model. 
For the first prompt, the model correctly labels the statement as misogynistic and also provides reasoning for its decision.

%Figure \ref{2} depicts the result of the second prompt given by the model. 
For the second prompt, the model assigns the following labels for the comment: `Stereotype', `Minimization and Trivialization', and `Victim Blaming', which are three of the four labels assigned by the annotator. This suggests that one-shot promoting LLMs can be a viable option for data labelling in hate speech detection.  

%\begin{figure}
%\includegraphics[width=\textwidth]{figures/prompt1.png}
%\caption{Prompt-1} \label{1}
%\end{figure}

%\begin{figure}
%\includegraphics[width=\textwidth]{figures/prompt2.png}
%\caption{Prompt-2} \label{2}
%\end{figure}

\section{Discussion} \label{discuss}

In this section, we will discuss the findings of the study and try to answer the research questions.

\begin{enumerate}
    \item How does zero-shot learning perform at the task of labelling YouTube comments for coarse and fine-grained misogyny classification in mix-code Hinglish?

    For binary classification, the zero-shot classifier achieved the highest accuracy of 54\%, which is not very high. For multi-label zero shot classification, the results are again very low. Previous studies indicated that the data labelling process for hate speech detection is a complex process \cite{basile2019semeval}. Different individuals might have a different concept of what constitutes hate speech, because of their varied socio-economic and cultural backgrounds \cite{waseem2016you} \cite{schmidt2017survey}. Also, the best results obtained by hate speech classifiers for low-resource and mix-code languages remain low, especially for fine grained classification \cite{mandl2020overview} \cite{fersini2018ibereval}. Therefore, a model that is not pre-trained on domain-specific data may have more difficulty in performing classification. Furthermore, the base classes of the comments have only been assigned by only one annotator and may contain some bias. The dataset will be additionally annotated by another annotator in future.

    \item How does one-shot and few-shot learning perform at the task of labelling YouTube comments for coarse and fine-grained misogyny classification in mix-code Hinglish?
    One-shot and few-shot misogyny classification was performed in this study using sentence transformers. Out of the two, few-shot classification achieves the higher scores on all the metrics. Macro-F1 score of the one-shot approach was 0.3571. which is significantly lower than 0.5108 for few-shot classification. The one-shot promoting results obtained through ChatGPT-3 are promising. Through the careful fine-tuning of prompts, the model correctly predicted whether the statement was misogynistic or not, and also predicted most of the fine-grained labels. This aligns with the findings of previous studies \cite{chiu2021detecting} \cite{del2023respectful}, which have concluded that LLMs perform well at the task of few-shot learning. 
    
    \end{enumerate}

Based on the findings of the study, one-shot prompting performs well on text labelling tasks. Out of zero-shot, one-shot, and few-shot classification, zero-shot performs the best. The results were not very good, but the study provides us with some directive about the future scope of using LLMs for data annotation for hate speech detection. Although the size of the dataset is very small, we received some indication that transfer learning can be beneficial in data annotation.

\section{Limitations} \label{limitations}

There a several limitations to the proposed study. Firstly, an LLM trained specifically trained on Hinglish has not been used to perform classification. Also, the labels used as the base value for comparison have only been weakly annotated by a single person, and they may have labelled the dataset according to their biases and worldview. The dataset is also small, consisting of only 100 comments. This study only examines 3 language models for zero and few-shot classification with their default parameters and does not provide any comparison with benchmarks.

\section{Conclusion and Future Work} \label{conc}

The field of hate speech detection requires innovation to be able to properly monitor offensive and hateful content. LLMs have been shown to learn a lot of information during the pre-training process and can be used for transfer learning. In this study, zero-shot, one-shot, and few-shot learning were used to label data for a weakly annotated dataset for misogyny detection. The results suggest that out of all the  techniques, zero-shot learning, and one-shot prompting techniques perform the best, implying that transfer learning can prove useful for data annotation. However, the empirical results for the zero-shot classification only reached 54\%. This could be accounted to the complexity of hate speech, biased labelling from a single annotator, and size of the dataset. Future work should focus on different approaches for zero-shot and few-shot techniques on a larger dataset to confirm our hypothesis. Also, rephrasing of the hypothesis for inference could be done to improve classification. Multilingual LLMs will also be explored for misogyny classification.

\bibliographystyle{splncs04}
\bibliography{ref.bib}

\end{document}